\documentclass[conference]{IEEEtran}
\IEEEoverridecommandlockouts
\usepackage{subfigure}

\usepackage{pbalance}

\IEEEoverridecommandlockouts 
\usepackage[dvips]{graphicx}
\usepackage{algorithmic}
\usepackage{algorithm}
\usepackage[OT4,T1]{fontenc}
\usepackage[cmex10]{amsmath}
\usepackage{url}

\usepackage{multirow}
\usepackage{hyperref}

\title{Passage Retrieval of Polish Texts Using OKAPI BM25 and an Ensemble of Cross Encoders}
\author{
\IEEEauthorblockN{Jakub Pokrywka}
\IEEEauthorblockA{
Adam Mickiewicz University\\
Faculty of Mathematics and Computer Science,\\
Email: jakub.pokrywka@amu.edu.pl}
}

\begin{document}
\maketitle              

\begin{abstract}
Passage Retrieval has traditionally relied on lexical methods like TF-IDF and BM25. Recently, some neural network models have surpassed these methods in performance. However, these models face challenges, such as the need for large annotated datasets and adapting to new domains. This paper presents a winning solution to the Poleval 2023 Task 3: Passage Retrieval challenge, which involves retrieving passages of Polish texts in three domains: trivia, legal, and customer support. However, only the trivia domain was used for training and development data. The method used the OKAPI BM25 algorithm to retrieve documents and an ensemble of publicly available multilingual Cross Encoders for Reranking. Fine-tuning the reranker models slightly improved performance but only in the training domain, while it worsened in other domains.
\end{abstract}

\section{Introduction}

Passage retrieval involves the task of retrieving a set of relevant text passages from a large collection of documents based on a given query. Typically, these passages are presented in descending order of relevance. The most commonly used method for passage retrieval is through lexical approaches like OKAPI BM25. Though, lexical models cannot capture semantic relationships between words, phrases, and sentences. To address this, neural language models can be employed. These models are often pretrained on extensive text corpora and then fine-tuned specifically for passage retrieval. There are two common setups for utilizing neural models in this task: complete passage retrieval using a neural model or combining another retrieval engine to retrieve a subset of passages, followed by using the neural model to select the most relevant ones. The latter approach is employed when the reranking model is too slow to process an entire document collection.

The Poleval 2023 Task 3: Passage Retrieval challenge aims to identify the best method for passage retrieval in Polish texts. The competition's test dataset comprises three domains: wiki-trivia, legal-questions, and allegro-faq. However, only the wiki-trivia domain is provided as the training and development dataset.


In this paper, we discuss the two-stage approach that achieved a score of 69.36 NDCG@10 on the final test competition dataset. Our method involves two phases. Firstly, we use the OKAPI BM25 algorithm to retrieve relevant passages. Then, an ensemble of Cross Encoder models is employed to rerank these passages. These models are publicly available multilingual models that have been trained on various languages (including Polish) and finetuned on multilingual corpora for passage reranking, as outlined in \cite{mMARCO}. We used these models with no further finetuning on the challenge dataset for two domains: legal-questions and allegro-faq. For the wiki-trivia domain, one model was fine-tuned and used in combination with models that had no further finetuning.

\begin{table}[h]
    \centering
    \caption{Dataset statistics split into given domains. rel. passages stand for relevant passages.} 
\begin{tabular}{c|c|c|c}
    - &  wiki-trivia & legal-questions & allegro-faq  \\
    \hline
    train questions & 4401 & 0 & 0  \\
    dev questions & 599 & 0 & 0 \\
    test-A questions & 400 & 400 & 400 \\
    mean test-A rel. passages & 3.46 & 1.97 & 1.09  \\
    test-B questions & 891 & 318 & 500 \\
    mean test-B rel. passages & 3.39 & 2.03 & 1.05\\
    passages & 7097322 &26287 & 921 \\
    mean word per passage & 44.6 & 155.1 & 50.0  \\
    
\end{tabular}
    \label{tab:datasetstats}
\end{table}
\section{Related Work}


\subsection{Reranker models and modern neural Information Retrieval}
MS MARCO \cite{msmarco} is a large publicly available reranking dataset retrieved by Bing. The dataset includes queries, retrieved documents by search engine, and a label on whether a user clicked a document. The corpus is in the English language. Recently, authors of mMARCO \cite{mMARCO} translated this corpus into many languages (but not into Polish though) and trained Cross Encoder reranker models on it. The base models were multilingual. The performance was effective not only for translated languages but also for not translated languages, only visible by models in the semisupervised pretraining phase.

BEIR \cite{beir} is an Information Retrieval benchmark for Zero-shot Evaluation between different domains. The authors provided many comparisons between different retrieval architectures. Very recently, the benchmark for Polish Information Retrieval was released in BEIR-PL paper \cite{beirpl}.

\subsection{Language models working on Polish texts}
There are a few transformer language models trained for the Polish language: HerBERT \cite{herbert}, plt5 \cite{plt5}, Polish RoBERTa \cite{polishroberta}. There are also many multilingual language models working on Polish languages, such as XLM-RoBERTa \cite{xlmroberta}, multilingual DeBERTa \cite{deberta}, and mT5 \cite{mt5}.


\section{Poleval 2023 Task 3: Passage Retrieval Challenge}

\subsection{Data}
The task is to retrieve the relevant passages given a query. The queries and passages are in the Polish language. There are separate domains:
wiki-trivia, legal-questions, allegro-faq. In the below subsection, each domain is presented. There are the following datasets: training (train), development (dev), test-A (preliminary test set), and test-B (final test set). For the training and development dataset, golden truth data was released during the competition, but the golden truth dataset was not. After competitions, the test set golden truth was released to \url{https://github.com/poleval/2022-passage-retrieval-secret}. Training and development datasets consist of only wiki-trivia, but the test dataset consists of all three domains. Below all domains are described. Some dataset statistics are given in Table \ref{tab:datasetstats}. Domains vary greatly in the number of passages and mean relevant passages per query.


\subsubsection{wiki-trivia}
Questions are general-knowledge typical for TV quiz shows, such as \textit{Fifteen to One} or Polish equivalent \textit{Jeden z dziesięciu}. For each question, there were manually selected up to five relevant passages (the mean number for the training dataset is 3.28 with a standard deviation of 1.45). The passages corpus consists of 7097322 elements. This domain was selected for train, dev, and test datasets. There are 4041 questions in the train dataset, 599 in the dev dataset, 400 in the test-A dataset, and 891 in the test-B dataset. Below, one example question with all correct passages is presented. \\ 

\textbf{Example Question:} "Czy w państwach starożytnych powoływani byli posłowie i poselstwa?" \\ 
\textbf{Correct passage:} "Poselstwo do Chin. Chińska "Księga Późniejszych Hanów" ("Houhanshu") zanotowała informację, iż w roku 166 drogą morską przez Ocean Indyjski do kraju Jinan (Annam) przybyło poselstwo "króla Andun z Daqin" (Rzymu), oferując kość słoniową, rogi nosorożca i szylkret." \\
\textbf{Correct passage:} "Z okresu starożytnych Indii pochodzą pierwsze zachowane instrukcje na temat zadań dyplomaty (rozpoznawać i informować) oraz cech dyplomaty (wykształcony, zręczny, zjednujący sobie ludzi). Dyplomację stosowali już starożytni Grecy, od których wywodzi się termin „diplóos” oznaczający złożony we dwoje dokument – rodzaj listów uwierzytelniających w które wyposażany był poseł." \\
\textbf{Correct passage:} "Ze starożytnej Grecji pochodzi również przywilej nietykalności posła i poselstwa (immunitet), które już wtedy były uznawane za zasadę będącą elementem prawa narodów. Pierwotną formą quasi-dyplomacji była wymiana posłów przez społeczności plemienne w celu przekazania i wyjaśnienia przekazu mocodawcy, lub wynegocjowania jakiegoś porozumienia (np. o zakończeniu wojny)."
\subsubsection{legal-questions} A portion of the legal questions were generated by randomly selecting provisions and formulating questions based on their content. The task is similar to SQuAD and requires only identifying relevant passages rather than answering the question.  The questions were supplemented with  26287 provisions derived from over one thousand laws published between 1993 and 2004. There are 400 questions in the test-A dataset and 318 in the test-B dataset. Below, the example questions from the test-A dataset are provided. 

\textbf{Example Question:} "Ile trwa kadencja szefa służby cywilnej?" \\
\textbf{Correct passage:} "Ustawa z dnia 18 grudnia 1998 r. o służbie cywilnej Rozdział 1 Przepisy ogólne","text":"Art. 9. 1. Szefa Służby Cywilnej powołuje, po zasięgnięciu opinii Rady Służby Cywilnej, Prezes Rady Ministrów spośró
d urzędników służby cywilnej. 2. Kadencja Szefa Służby Cywilnej trwa 5 lat, licząc od dnia powołania; Szef Służby Cywilnej pełni obowiązki do dnia powołania jego następcy. 3. Kadencja Szefa Służby Cywilnej wygasa w razie jego śmierci lub odwołania. 4. Prezes Rady Ministrów odwołuje Szefa Służby Cywilnej w razie: 1) rezygnacji ze stanowiska, 2) utraty zdolności do pełnienia powierzonych obowiązków na skutek długotrwałej choroby, trwającej co najmniej 6 miesięcy. 5. Prezes Rady Ministrów odwołuje Szefa Służby Cywilnej także w przypadku, gdy przestał on odpowiadać jednemu z warunków określonych w art. 4. Odwołanie w przypadkach, o których mowa w art. 4 pkt 4 i 5, następuje za zgodą co najmniej 23 składu Rady Służby Cywilnej."
\subsubsection{allegro-faq} Questions regards the large e-commerce platform- Allegro.pl were created using help articles and lists of frequently asked questions. There are 400 questions in the test-A dataset and 500 questions in the test-B dataset. In total, there are 921 passages. Here is an example question from the test-A dataset:  

\textbf{Example Question:} "Otrzymałem rekompensatę z POK, a później zwrot od Sprzedającego. Co mam zrobić?"  \\
\textbf{Correct passage:} "Jeśli wypłaciliśmy Ci rekompensatę w ramach Programu Ochrony Kupujących a Ty otrzymasz zwrot pieniędzy od Sprzedającego, masz obowiązek zwrócić nam rekompensatę. Zgodnie z punktem 6 Część IV Załącznika nr 9 do Regulaminu Allegro, na zwrot rekompensaty masz 7 dni od naprawienia szkody przez Sprzedającego."


\subsection{Evaluation Metric}

The Geval evaluation tool \cite{geval} uses Normalized Discounted Cumulative Gain for the top ten passages (NDCG@10) as the challenge metric. The challenge was hosted on the Gonito platform \cite{gonito}, and the final evaluation was conducted on the test-B dataset across all domains. It should be noted that the sample split between domains is not equal, which means that some domains have a greater impact on the final score.


\section{Method}

The solution involves two stages: Retrieval and Reranking. Retrieval is carried out using the lexical method OKAPI BM25, which is quick but not as effective as a neural ranking model. Additionally, it does not require training. The best performing method for Reranking is through Cross Encoders, but it is slow as it requires processing every query-passage pair. Due to its time-consuming nature, it can only operate on a limited set of passages, except for the allegro-faq domain, which consists of only 921 passages.
\subsection{Retrieval phase}
For retrieval model We used OKAPI BM25 algorithm using parameters $k_1$=1.2 , $b$=0.75, $\varepsilon$=0.25. The utilized library may be accessed via \url{https://github.com/zhusleep/fastbm25}. The preprocessing included tokenization using the nltk library, specifically nltk.tokenize.word\_tokenize, lowercase normalization, stemming using pystempel (accessed via \url{https://github.com/dzieciou/pystempel} ) with the Polimorf \cite{polimorf} stemmer, and removal of Polish stopwords.

\subsection{Reranking phase}

The reranking phase was performed using an ensemble of multilingual reranker models based on Cross-Encoder architecture. We used different sets of the ensemble for wiki-trivia domain and  legal-questions with allegro-faq questions. Both are described in the following section. The ensembles were created by summing up all the individual models' probability scores. Finetuning, if performed, was loosely based on a script from Sentence-Transformer library \cite{sentencebert}, namely \url{https://github.com/UKPLab/sentence-transformers/blob/master/examples/training/ms_marco/train_cross-encoder_scratch.py}. The process of finetuning and inference was completed on A100 GPU card. We used one 100 negative query-passage pair for each positive passage selected from the training dataset. The negative passage selection was from the top 2000 passages returned by the described OKAPI BM25 algorithm. The used Loss was BCEWithLogitsLoss with a constant learning rate scheduler of 1e-6 and 2000 warmup steps. The best-performing model was selected for inference from training for ten epochs.

\subsubsection{wiki-trivia}
Reranking was based on the top 3000 results from the OKAPI BM25 algorithm. Because wiki-trivia passages are relatively short, they only require a little time, although, during experiments, we observed that reranking with above 1000 passages, there is not much gain in the metric score.

The ensemble consisted of three models:

\begin{itemize}
    \item Publicly available reranker based on multilingual T5 (mT5) model \cite{mt5} (also trained on Polish corpora) and fine-tuned to automatically translated reranking corpus MS MARCO \cite{msmarco} into Portuguese. The model unicamp-dl/mt5-13b-mmarco-100k via \url{https://huggingface.co/unicamp-dl/mt5-13b-mmarco-100k} was used as it is without further fine-tuning to the competition training dataset. Therefore model works in a zero-shot manner as described in \cite{mMARCO}.
    \item Reranker cross-encoder/mmarco-mMiniLMv2-L12-H384-v1 ( accesed via \url{https://huggingface.co/cross-encoder/mmarco-mMiniLMv2-L12-H384-v1}. The multilingual base model MiniLMv2 \cite{minilmv2} is fine-tuned on mMARCO dataset (MS MARCO translated into multiple languages). Please note MMARCO dataset does not contain the Polish language, but MiniLMv2 was trained on Polish. However, it performed well on the dev dataset. We then fine-tuned it further on the competition train dataset, slightly improving it.
    \item Reranker cross-encoder/mmarco-mdeberta-v3-base-5negs-v1 (\url{https://huggingface.co/cross-encoder/mmarco-mdeberta-v3-base-5negs-v1}) based on multilingual DeBERTaV3 \cite{deberta} finetuned on MMARCO dataset. During the competition, the model was publicly available but was removed before the time of writing this article. We further fine-tuned the model to the competition training dataset.
\end{itemize}

\subsubsection{legal-questions and allegro-faq}

Reranking was performed on top 1500 passages for legal-questions. The limit was lower than for wiki-trivia due to the length of passages collection and longer computation time. For the allegro-faq domain, reranking was performed on all the passages since the whole collection consists of only 921 passages. For both domains, the same ensemble was used. The following models were used without further finetuning to the competition dataset. We conducted experiments using models fine-tuned to wiki-trivia, but their performance dropped drastically. Finally, we used the following models:
\begin{itemize}
    \item Model  unicamp-dl/mt5-13b-mmarco-100k via \url{https://huggingface.co/unicamp-dl/mt5-13b-mmarco-100k} described in the previous section.
    \item  Model unicamp-dl/mt5-3B-mmarco-en-pt via \url{https://huggingface.co/unicamp-dl/mt5-3B-mmarco-en-pt}, which is the same as above but in the 3B parameters version.
\end{itemize}


\section{Results}
The presented method scores 75.40 NDCG@10 on preliminary test-A and on 69.36 NDCG@10 on final test-B data. The experiments code is available at \url{https://github.com/kubapok/poleval22}. The analysis of single models on different reranking size limits is presented in Table \ref{tab:testaresults} for test-A and in Table \ref{tab:testbresults} for test-B. The results vary between domains, probably because of text nature, as well as different passage collection sizes and different size mean relevant passages per one query. All the presented reranking models score better than the OKAPI BM25 baseline. With the reranking size limit, the performance is better. However, the gain isn't great beyond the reranking limit of 500. Finetuning models increase their performance only on the wiki-trivia domain and worsen on other domains. Unfortunately, these results are not included in the presented tables as we didn't save them.
\section{Other Experiments}
We have tried other approaches as well. These experiments were very preliminary and may yield better results if we spend more time on them. However, we decided to include them in this paper anyway.

\begin{table*}[htp]
    \centering
    \caption{NDCG@10 results for the whole final testing dataset test-B and split into domains. ft stands for the model fine-tuning to the competition data, whereas no-ft stands for no fine-tuning. The number at the right of the model name stands for the reranking size from the OKAPI BM25 algorithm. Some experiments were not conducted or saved. In this case, the score is labeled as "-"} 
\begin{tabular}{c|c|c|c|c}
    model & test-B & wiki-trivia & legal-questions & allegro-faq  \\
    \hline
    final ensemble & 69.36  &  55.13 &  86.39 & 83.88   \\
    \hline
    OKAPI BM25 & 42.55 &23.48 &81.31 &51.87 \\
    \hline
    mmarco-mMiniLMv2-L12-H384-v1 no-ft 10  & 48.85 &28.45 &83.00 &63.47\\ 
    mt5-3B-mmarco no-ft  10 &  50.31 &29.47 &84.35 &65.81 \\ 
    mt5-13B-mmarco no-ft  10  & 50.36 &29.63 &83.59 &66.15 \\ 
    \hline
    mmarco-mMiniLMv2-L12-H384-v1 no-ft 50 & 56.18 &35.88 &85.26 &73.84 \\
    mt5-3B-mmarco no-ft  50 & 59.04 &38.06 &86.75 &78.80 \\ 
    mt5-13B-mmarco no-ft  50  &  59.79 &39.30 &85.30 &80.08 \\
    \hline
    mmarco-mMiniLMv2-L12-H384-v1 no-ft 100 & 57.76 &38.22 &85.54 &74.91 \\
    mt5-3B-mmarco no-ft  100 & 61.42 &41.24 &87.06 &81.09 \\ 
    mt5-13B-mmarco no-ft  100  & 62.65 &43.17 &85.63 &82.75 \\
    \hline
    mmarco-mMiniLMv2-L12-H384-v1 no-ft 500 & 58.52 &39.86 &85.61 &74.56 \\
    mt5-3B-mmarco no-ft  500 &  63.48 &44.41 &86.67 &82.70\\ 
    mt5-13B-mmarco no-ft  500  &  65.04 &47.21 &85.42 &83.86 \\
    \hline
    mmarco-mMiniLMv2-L12-H384-v1 no-ft 1000 & 58.91 &40.49 &85.66 &74.72 \\
    mt5-3B-mmarco no-ft  1000 & 64.12 &45.48 &86.64 &83.01 \\ 
    mt5-13B-mmarco no-ft  1000  &  65.59 &48.13 &85.22 &84.21 \\
    \hline
    mmarco-mMiniLMv2-L12-H384-v1 no-ft 1500 & 58.99 & 40.70 & 85.51 & 74.72 \\
    mmarco-mMiniLMv2-L12-H384-v1 ft 1500 & - & 47.64 & - & - \\
    mmarco-mdeberta-v3-base-5negs-v1 no-ft 1500 & - & 45.30 & - & - \\
    mmarco-mdeberta-v3-base-5negs-v1 ft 1500 & - & 51.73 & - & - \\
    mt5-3B-mmarco no-ft  1500 & 64.46 & 46.17 & 86.55 & 83.01  \\
    mt5-13B-mmarco no-ft  1500& 65.99 &  48.96 &  85.04 & 84.21  \\
\end{tabular}
    \label{tab:testbresults}

    \bigskip
    \centering
    \caption{NDCG@10 results for the whole preliminary testing dataset test-A and split into domains. ft stands for the model fine-tuning to the competition data, whereas no-ft stands for no fine-tuning. The number at the right of the model name stands for the reranking size from the OKAPI BM25 algorithm. Some experiments were not conducted or saved. In this case, the score is labeled as "-"} 
\begin{tabular}{c|c|c|c|c}
    model & test-A & wiki-trivia & legal-questions & allegro-faq  \\
    \hline
    final model & 75.40 & 52.25  & 86.48 & 87.48   \\
    \hline
    OKAPI BM25 & 52.67 &22.26 &81.78 &53.96 \\
    \hline
    mmarco-mMiniLMv2-L12-H384-v1 no-ft 10  & 58.81 &26.03 &84.95 &65.46 \\ 
    mmarco-mdeberta-v3-base-5negs-v1 no-ft 10 & 59.52 &26.69 &84.79 &67.09 \\
    mt5-base-mmarco-v2 no-ft 10 &  58.60 &25.91 &83.96 &65.95 \\
    mt5-3B-mmarco no-ft  10 & 60.14 &27.09 &84.74 &68.60 \\ 
    mt5-13B-mmarco no-ft  10  & 60.09 &27.30 &84.00 &68.98 \\ 
    \hline
    mmarco-mMiniLMv2-L12-H384-v1 no-ft 50 & 65.13 &32.81 &85.60 &76.98 \\
    mmarco-mdeberta-v3-base-5negs-v1 no-ft 50 &66.96 &35.17 &85.92 &79.80 \\
    mt5-base-mmarco-v2 no-ft 50 &  64.97 &33.14 &84.31 &77.45 \\
    mt5-3B-mmarco no-ft  50 &  68.24 &35.81 &85.57 &83.33 \\ 
    mt5-13B-mmarco no-ft  50  &  68.78 &36.70 &84.90 &84.75 \\
    \hline
    mmarco-mMiniLMv2-L12-H384-v1 no-ft 100 & 66.31 &35.32 &85.96 &77.66 \\
    mmarco-mdeberta-v3-base-5negs-v1 no-ft 100 &  68.39 &38.27 &86.28 &80.63 \\
    mt5-base-mmarco-v2 no-ft 100 & 65.70 &35.04 &84.35 &77.70 \\
    mt5-3B-mmarco no-ft  100 &69.97 &38.82 &86.10 &84.99 \\ 
    mt5-13B-mmarco no-ft  100  &  70.83 &40.43 &85.42 &86.63 \\
    \hline
    mmarco-mMiniLMv2-L12-H384-v1 no-ft 500 & 67.11 &37.46 &85.80 &78.05 \\
    mmarco-mdeberta-v3-base-5negs-v1 no-ft 500 & 69.31 &41.02 &85.92 &80.99 \\
    mt5-base-mmarco-v2 no-ft 500 & 65.85 &36.50 &83.81 &77.25 \\
    mt5-3B-mmarco no-ft  500 &  71.40 &42.13 &86.14 &85.93 \\ 
    mt5-13B-mmarco no-ft  500  & 72.45 &43.94 &85.50 &87.91 \\
    \hline
    mmarco-mMiniLMv2-L12-H384-v1 no-ft 1000 & 67.37 &38.29 &85.69 &78.11 \\
    mmarco-mdeberta-v3-base-5negs-v1 no-ft 1000 & 69.73 &42.30 &85.85 &81.04 \\
    mt5-base-mmarco-v2 no-ft 1000 &  65.98 &36.96 &83.66 &77.32 \\
    mt5-3B-mmarco no-ft  1000 & 71.84 &43.29 &86.20 &86.03 \\ 
    mt5-13B-mmarco no-ft  1000  &  73.06 &45.54 &85.66 &88.00 \\
    \hline
    mmarco-mMiniLMv2-L12-H384-v1 no-ft 1500 &  67.35& 38.45 &85.50 &78.11 \\
    mmarco-mMiniLMv2-L12-H384-v1 ft 1500 &  - & 45.84 &- &- \\
    mmarco-mdeberta-v3-base-5negs-v1 no-ft 1500 & 69.82 &42.58 &85.82 &81.04 \\
    mmarco-mdeberta-v3-base-5negs-v1 ft 1500 & - & 48.99 & - & - \\
    mt5-base-mmarco-v2 no-ft 1500 & 65.99 &37.11 &83.54 &77.32 \\
    mt5-3B-mmarco no-ft  1500 & 72.01 & 43.78 & 86.22 &  86.03  \\
    mt5-13B-mmarco no-ft  1500 & 73.28 & 46.26 &  85.57 & 88.00\\
\end{tabular}
    \label{tab:testaresults}
\end{table*}

\subsection{Translating Polish texts into English.}
We translated Polish passages and queries into English using a machine translation model accessed by \url{https://huggingface.co/gsarti/opus-mt-tc-en-pl} \cite{opus}. English Cross Encoder reranking models did not perform on the translated texts better than multilingual reranking models on Polish texts tough.

\subsection{Bi Encoder models}
We experimented with various publicly available Bi Encoder models, using them as one-stage retrieval models. Unfortunately, their performance was significantly inferior to that of the OKAPI BM25 algorithm operating alone. However, combining the OKAPI BM25 and Bi Encoder models as retrieval models for further reranking with the Cross Encoder model may lead to improved results and is a promising area for research. Our highest Bi Encoder score for untranslated documents was 9.26 NDCG@10, achieved using the sentence-transformers/distiluse-base-multilingual-cased-v1 model. For translated texts into English, our highest score was 21.00, obtained using the sentence-transformers/all-mpnet-base-v2 model.
\subsection{Translating MS MARCO into Polish}
MMARCO does not include translations for Polish texts. We've attempted translating MS MARCO into English using model gsarti/opus-mt-tc-en-pl and training several reranking models on this data. The approach is similar to \cite{beirpl}. Nevertheless, this work was published after the competition. In our case, this approach didn't yield better results than large multilingual models.

\section{Conclusions}
This paper summarizes our solution to Poleval 2023 Task 3: Passage Retrieval. The system operates in two stages, utilizing OKAPI BM25 for retrieval and a multilingual ensemble of Cross Encoders for reranking. However, the system's performance varies between domains due to the limited availability of training data for only one domain. While fine-tuning the neural model can enhance results for this domain, it may have a negative impact on other domains.
\bibliography{bibliography}

\begin{thebibliography}{10}

\bibitem{mMARCO}
L.~Bonifacio, V.~Jeronymo, H.~Q. Abonizio, I.~Campiotti, M.~Fadaee, R.~Lotufo,
  and R.~Nogueira, ``mmarco: A multilingual version of the ms marco passage
  ranking dataset,'' 2021.

\bibitem{msmarco}
T.~Nguyen, M.~Rosenberg, X.~Song, J.~Gao, S.~Tiwary, R.~Majumder, and L.~Deng,
  ``Ms marco: A human generated machine reading comprehension dataset.,'' {\em
  CoRR}, vol.~abs/1611.09268, 2016.

\bibitem{beir}
N.~Thakur, N.~Reimers, A.~R{\"u}ckl{\'e}, A.~Srivastava, and I.~Gurevych,
  ``{BEIR}: A heterogeneous benchmark for zero-shot evaluation of information
  retrieval models,'' in {\em Thirty-fifth Conference on Neural Information
  Processing Systems Datasets and Benchmarks Track (Round 2)}, 2021.

\bibitem{beirpl}
K.~Wojtasik, V.~Shishkin, K.~Wo{\l}owiec, A.~Janz, and M.~Piasecki, ``Beir-pl:
  Zero shot information retrieval benchmark for the polish language,'' {\em
  arXiv preprint arXiv:2305.19840}, 2023.

\bibitem{herbert}
R.~Mroczkowski, P.~Rybak, A.~Wróblewska, and I.~Gawlik, ``{H}er{BERT}:
  Efficiently pretrained transformer-based language model for {P}olish,'' in
  {\em Proceedings of the 8th Workshop on Balto-Slavic Natural Language
  Processing}, (Kiyv, Ukraine), pp.~1--10, Association for Computational
  Linguistics, Apr. 2021.

\bibitem{plt5}
A.~Chrabrowa, {\L}.~Dragan, K.~Grzegorczyk, D.~Kajtoch, M.~Koszowski,
  R.~Mroczkowski, and P.~Rybak, ``Evaluation of transfer learning for polish
  with a text-to-text model,'' {\em arXiv preprint arXiv:2205.08808}, 2022.

\bibitem{polishroberta}
S.~Dadas, M.~Pere{\l}kiewicz, and R.~Po{\'{s}}wiata, ``Pre-training polish
  transformer-based language models at scale,'' in {\em Artificial Intelligence
  and Soft Computing}, pp.~301--314, Springer International Publishing, 2020.

\bibitem{xlmroberta}
A.~Conneau, K.~Khandelwal, N.~Goyal, V.~Chaudhary, G.~Wenzek, F.~Guzm{\'{a}}n,
  E.~Grave, M.~Ott, L.~Zettlemoyer, and V.~Stoyanov, ``Unsupervised
  cross-lingual representation learning at scale,'' {\em CoRR},
  vol.~abs/1911.02116, 2019.

\bibitem{deberta}
P.~He, X.~Liu, J.~Gao, and W.~Chen, ``Deberta: Decoding-enhanced bert with
  disentangled attention,'' in {\em International Conference on Learning
  Representations}, 2021.

\bibitem{mt5}
L.~Xue, N.~Constant, A.~Roberts, M.~Kale, R.~Al-Rfou, A.~Siddhant, A.~Barua,
  and C.~Raffel, ``m{T}5: A massively multilingual pre-trained text-to-text
  transformer,'' in {\em Proceedings of the 2021 Conference of the North
  American Chapter of the Association for Computational Linguistics: Human
  Language Technologies}, (Online), pp.~483--498, Association for Computational
  Linguistics, June 2021.

\bibitem{geval}
F.~Grali{\'n}ski, A.~Wr{\'o}blewska, T.~Stanis{\l}awek, K.~Grabowski, and
  T.~G{\'o}recki, ``{GE}val: Tool for debugging {NLP} datasets and models,'' in
  {\em Proceedings of the 2019 ACL Workshop BlackboxNLP: Analyzing and
  Interpreting Neural Networks for NLP}, (Florence, Italy), pp.~254--262,
  Association for Computational Linguistics, Aug. 2019.

\bibitem{gonito}
F.~Grali{\'n}ski, R.~Jaworski, {\L}.~Borchmann, and P.~Wierzcho{\'n},
  ``Gonito.net -- open platform for research competition, cooperation and
  reproducibility,'' in {\em Proceedings of the 4REAL Workshop: Workshop on
  Research Results Reproducibility and Resources Citation in Science and
  Technology of Language} (A.~Branco, N.~Calzolari, and K.~Choukri, eds.),
  pp.~13--20, 2016.

\bibitem{polimorf}
W.~Kieraś and M.~Woliński, ``Morfeusz 2 – analizator i~generator fleksyjny
  dla języka polskiego,'' {\em Język Polski}, vol.~XCVII, no.~1, pp.~75--83,
  2017.

\bibitem{sentencebert}
N.~Reimers and I.~Gurevych, ``Making monolingual sentence embeddings
  multilingual using knowledge distillation,'' in {\em Proceedings of the 2020
  Conference on Empirical Methods in Natural Language Processing}, Association
  for Computational Linguistics, 11 2020.

\bibitem{minilmv2}
W.~Wang, H.~Bao, S.~Huang, L.~Dong, and F.~Wei, ``{M}ini{LM}v2: Multi-head
  self-attention relation distillation for compressing pretrained
  transformers,'' in {\em Findings of the Association for Computational
  Linguistics: ACL-IJCNLP 2021}, (Online), pp.~2140--2151, Association for
  Computational Linguistics, Aug. 2021.

\bibitem{opus}
J.~Tiedemann and S.~Thottingal, ``{OPUS-MT} — {B}uilding open translation
  services for the {W}orld,'' in {\em Proceedings of the 22nd Annual Conferenec
  of the European Association for Machine Translation (EAMT)}, 2020.

\end{thebibliography}
\bibliographystyle{ieeetr}
\end{document}